# Landau Theory of Adaptive Integration in Computational Intelligence

*Dariusz Plewczynski* [1*]

[1] Interdisciplinary Centre for Mathematical and Computational Modelling, University of Warsaw,

Pawinskiego 5a Street, 02-106 Warsaw, Poland, E-mail: darman@icm.edu.pl

[*] To whom correspondence should be addressed

**Abstract**

*Computational Intelligence* (CI) is a sub-branch of Artificial Intelligence paradigm focusing on the study of adaptive mechanisms to enable or facilitate intelligent behavior in complex and changing environments. There are several paradigms of CI [like artificial neural networks, evolutionary computations, swarm intelligence, artificial immune systems, fuzzy systems and many others], each of these has its origins in biological systems [biological neural systems, natural Darwinian evolution, social behavior, immune system, interactions of organisms with their environment]. Most of those paradigms evolved into separate machine learning (ML) techniques, where probabilistic methods are used complementary with CI techniques in order to effectively combine elements of learning, adaptation, evolution and Fuzzy logic to create heuristic algorithms that are, in some sense, intelligent. The current trend is to develop consensus techniques, since no single machine learning algorithms is superior to others in all possible situations. In order to overcome this problem several meta-approaches were proposed in ML focusing on the integration of results from different methods into single prediction. We discuss here the Landau theory for the nonlinear equation that can describe the adaptive integration of information acquired from an ensemble of independent learning agents. The influence of each individual agent on other learners is described similarly to the social impact theory. The final decision outcome for the consensus system is calculated using majority rule in the stationary limit, yet the minority solutions can survive inside the majority population as the complex intermittent clusters of opposite opinion.

## Introduction

Several meta-approaches were proposed in machine learning (ML) field focusing on the integration of results from different algorithms into single prediction [1-9]. The typical meta-learning procedure is trying to balance the generality of solution and the overall performance of trained model. The main problem with such meta-approaches is that they are static, i.e. no adaptivity is included. The meta-approach is typically optimized for certain combination of single machine learning methods and particular representation of training data. Yet, the actual output of the training should impact the parameters of the method, then allowing for iterative procedure that is able to adapt to changing environment and further optimization of training model. This dynamical view of machine learning is especially useful for robotic vision applications [10, 11], general robots [10, 12-20], and bioinformatics [6, 21-24].

The term *Adaptive Integration* (AI) describes this adaptive behavior, when the integration of results from ensemble of ML methods is impacted by its output, as it was applied in various approaches in bioinformatics. Here, the balance between environment and trained model can be described similarly to social influence theory as the global parameter affecting all learners. Such methodology is directly taken from *Computational Intelligence* (CI) [25], where each intelligent agent performs training on available input data toward classification pressure described by the set of positive and negative cases. When the query testing data is analyzed each agent predicts the query item classification by "yes"/"no" decision. The answers of all agents are then gathered and fused into the single prediction. The integration scheme allows for adaptive changes when different set of input data is presented to the system by retraining all learners.

The first mathematical approach to the analysis of opinion formation in groups of individuals was made by Abelson [26]. It was proven that a wide class of linear models of individual attitude change lead to uniformity of opinions, which is not a case of most real-world phenomena. Further work of Axelrod et al. has led to a deeper understanding of the self-organization of minority and majority and other factors in the evolution of cooperation: the number of agents, the range of possible choices, noise, population dynamics together with population structure [27-31]. The other class of models that is based on probabilistic cellular automata was later proposed by Nowak et al. [32]. They studied a computer simulation model of the change of attitudes in a population resulting from the interactive, reciprocal, and recursive operation of previously known Latane's theory of social impact [33]. Surprisingly, several emerging macroscopic phenomena were observed, yet resulting from relative simple operation of microscopic rules of opinion change. They observed an incomplete polarization of opinions reaching a stable equilibrium, with coherent minority subgroups managing to exist near the margins of the whole population [32].

The mean-field theory with intermittent behavior shows a variety of stationary states with a well-localized and dynamically stable clusters (domains) of individuals who share minority opinions. The statistical mechanical model of social impact formulated by Latane [33] and then extended by Lewenstein [34] postulates the impact of a group of $N$ agents on a given learner is proportional to three factors: the *"strength"* of the members of the whole ensemble, their *"social"* distance from the individual, and their number $N$. These model leads to ferromagnetic and spin-glass phases, when different values of persuasiveness and supportiveness are assumed. Later, the extension of the model was done by Kohring [35, 36], where Latane's theory was extended to include learning. The Ising model of social impact allows for more detailed analysis of the impact of connections strengths on the final opinion clusters. The Lewenstein's class of models of cellular automata with intrinsic disorder was later extended to continuous limit by Plewczynski [37], and proved that even the model of Cartesian social space (therefore not fully connected) and containing no learning rules, one can also observe different phases (small clusters in the sparse phase with large role of strong individuals, and high density phase with almost uniform opinion).

Similar model of opinion formation based on the theory of social impact and the concept of cellular automata was later analyzed numerically in the special case of a strong leader and an external social impact acting uniformly on every individual [38]. They found two basic stationary states of the system: cluster of the leader's adherents and unification of opinions. The variation of model parameters (leader's strength or external impact) in the deterministic limit changes the size of the cluster or even forces the system to jump to another global solution. They analyzed by numerical simulations also the influence of noise by proposing transitions like inverses (flips) of the unified group opinion due to random flips of the leader's opinion [38]. Further work of Hołyst et al. focused on more detailed analysis of phase transitions in such simplified social impact theory models with single leader or in the case of Brownian particle representation of individuals [39]. Similar rapid changes of the opinion distribution with a continuous change of a system parameter are observed in random connections topology of social space with different probability distributions [40]. The scale-free network as the social space for a system was later modeled numerically by Grabowski et al. They also observed critical phenomena described by the critical temperature of the system $TC$ or by global parameter weighting the influence of external stimuli on the social network (in that case a mass media influence on the whole population) [41, 42].

The paper is organized as follows. In section 2 we describe the discrete model of learning, which is then extended to continuous case in section 3. The section 4 is devoted to the formulation of mean-field equation of dynamics and the Lyapunov functional of the system. In the next section the stationary solutions and the phases in the model are described. The section 6 contains the model of equilibration of the system, and the last part of the article elucidates the main outcomes of my work.

**Cellular Automata model of meta-learning**

Our model of learning is an application of nonlocal cellular automata approach known from physics. The original model in the context of social opinion exchange was formulated by Lewenstein et al. [34], where individual differences of all subsystems and the *social* influence decaying with distance are assumed. Within this general statistical model the integration of opinion between individual learners (agents) is similar to the dynamical approach of the statistical system to its stationary state. Therefore the phase transitions can be observed in the system, where the whole system reaches locally a phase change point, emerging the global solution. In our case, the global solution is either uniform, or presenting a variety of clustered minority opinions, within the global sea of majority. This is described here as *integration* of all learning models into the single consensus solution. Changes between phases of the system are induced by some external factors that can be modeled as a bias added to the local fields (minority clusters are free to grow, when they agree with their opinion with the bias value). In that way, the global opinion change can be easily modeled while the external influence changes giving rise to new, global majority in the learning system. This explains the *adaptivity* of the learning algorithm, when the system dynamically responds on the change in the training or input data, allowing for rapid adaptation of the final prediction (for example classification outcome). Such adaptivity effect describes the memory of the system, where it preserves the previous solutions not impacting the adaptation to emerged new training data.

The model of meta-learning is based on several assumptions:

1. *Binary Logic*

   We assume binary logic of individual learners, i.e. we deal with cellular automata consisting of $N$ agents, each holding one of two opposite states ("No" or "YES"). These states are binary $\sigma_i = \pm 1$, similarly to Ising model of ferromagnet. In most cases the machine learning algorithms that can model those agents, such as support vector machines, decision trees, trend vectors, artificial neural networks, random forest, predict two classes for incoming data, based on previous experience in the form of trained models. The prediction of an agent answers single question: is a query data contained in class A ("YES"), or it is different from items gathered in this class ("NO").

2. *Disorder and random strength parameter*

   Each learner is characterized by two random parameters: persuasiveness $p_i$ and supportiveness $s_i$ that describe how individual agent interact with others. Persuasiveness describes how effectively the individual state of agent is propagated to neighboring agents, whereas supportiveness represent self-supportiveness of single agent. In present work we assume that influential agents has the high self-esteem $p_i = s_i$, what is supported by the fact that highly effective learners should have high impact on others in the meta-learning procedure. In general the individual differences between agents are described as random variables with a probability density $\hat{p} = (p_i, s_i)$. Similarly to the social influence theory, the quality of predictor in some way affects its influence strength, when the final optimization of the meta-learning consensus is done.

3. *Learning space and learning metric*

   Each agent is characterized by its location in the learning space, therefore one can calculate the abstract learning distance $d(i, j)$ of two learners *i* and *j*. The strength of coupling between two agents tends to decrease with the learning distance between them. Determination of the learning metric is a separate problem, and the particular form of the metric and the learning distance function should be empirically determined, and in principle can have a very peculiar geometry. For example the learning space can be represented as a linear vector of learning agents, with distance between neighbors equal to 1 (used in short functional motifs identification in bioinformatics), two or three dimensional matrix of agents with Euclidean distance metric that can be used in computer vision problems. In the case of simple consensus between different machine learning algorithms the fully connected learning space has to be introduced, where all distances between agents are equal. The hierarchical geometry with ultrametric distance describes the splitting of an ensemble of learners into hierarchical groups, capturing different features or scales in the training data. There, the distances between agents are determined by their hierarchy level. In the case of strongly diluted model, agents are randomly connected to others without any metric structure. In our present manuscript we will analyze two dimensional Euclidian geometry, with a constant, finite-range metric. In that case, the decay of learning coupling is described by a function $1/g(d(i,j))$, equal to constant value $g$ for $d(i,j) < R$, and $\infty$ for more distant pairs. In addition we choose $g(0) = 1/\beta$, where $\beta = 1/kT$, and $T$ represent temperature of the system, that allows for simulating the competition between persuasiveness and supportiveness of each agent.

4. *Learning coupling*

   Agents exchange their opinions by biasing others toward their own classification outcome. This influence can be described by the total learning impact $I_i$ that *i*th agent is experiencing from all other learners. Within the cellular automata approach this impact is the difference between positive coupling of those agents that hold identical classification outcome, relative to negative influence of those who share the opposite state, and can be formalized as

$$I_i = I_p\left(\sum_j \frac{t(p_j)}{g(d(i,j))}(1-\sigma_i\sigma_j)\right) - I_s\left(\sum_j \frac{t(s_j)}{g(d(i,j))}(1+\sigma_i\sigma_j)\right), \tag{1}$$

where $g(d(i,j))$ is a decreasing function of distance $d(i,j)$, and $t(p_i,s_j)$ is the strength scaling function. The strength scaling can be taken to be $t(x)=x$, providing redefinition of the probability density distribution $\hat{p}=(p_i,s_i)$.

The equation of dynamics of the learning model defines the state $\sigma_i^{'}$ of *i*th individual at the next time step as follows:

$$\sigma_i^{'} = (-sign(\sigma_i I_i)), \tag{2}$$

with rescaled learning influence:

$$I_i = \sum_j \frac{p_j}{(s+p)g(d(i,j))}(1-\sigma_i\sigma_j) - \sum_j \frac{s_j}{(s+p)g(d(i,j))}(1+\sigma_i\sigma_j). \tag{3}$$

We assume a synchronous dynamics, i.e. states of all agents are updated in parallel. In comparison to standard Monte Carlo methods the synchronous dynamics takes shorter time to equilibrate than serial methods, yet it can be trapped into periodic asymptotic states with oscillations between neighboring agents.

5. Presence of *noise*

   The randomness of the state change (phenomenological modeling of various random elements in the learning system, and training data) is given by introducing a noise into the dynamics:

$$\sigma_i^{'} = (-sign(\sigma_i I_i + h_i)) \tag{4}$$

   where $h_i$ is the site-dependent white noise, or one can select a uniform white noise, where for all agents $h_i = h$. In the first case, $h_i$ are random variables independent for different agents and time points, whereas in the second case *h* are independent for different time points. We assume here, that the probability distribution of $h_i$ is both site and time independent, i.e. it has uniform statistical properties. The uniform white noise simulates the global bias affecting all agents (like impurities in training data), whereas site-dependent white noise describes local effects (such as prediction quality of individual learner etc.).

The system defined in this way is similar to previously postulated cellular automata models of opinion change in social sciences [34, 37]. The main differences of those approaches from the previously described cellular automata models is given by the short-range interactions. In addition, the random strength parameters are introduced, therefore allowing for more complex behavior of the system. Individual agents are described using probability density $\hat{p}=\left(p_i,s_i\right)$, so they differs from each other. The impact function is also included, so learners are able to exchange their states by the coupling procedure. The optimization goal of the meta-learning procedure is constructed here by minimizing the free energy of the system with the constrains imposed by the existence of the single solution, and that the probabilities of both answers should sum to one.

**Mean-Field Model for Euclidean Geometry**

The *n*-dimensional learning space geometry presents an interesting real-life case for further analytical analysis. There, the coupling between agents decreases with increasing Euclidean distance between them. The mean-field theory provides very well defined and controlled approximation allowing for solving the dynamical equations of the model. The dynamical "order" parameter has to be defined, to show the decay of minority groups in the form of "staircase" dynamics [34, 37, 43, 44]. The *n*-dimensional geometry of learning space is supported for example by the topology of connections in a mammalian brain. Here, the sub-regions (like vision cortex) have approximately three-dimensional geometry, sometimes with complex boundary conditions (for example cylindrical ones). In that case, a network of neurons is integrating information via non-trivial couplings between individual neurons, and the final stationary phase is determined by both topology of connections, and the strength of couplings between agents. Yet, this process can be locally described as the nearest-neighbor coupling in three dimensional Euclidean space, if the mean size of information integration region is smaller than the mean size of the selected sub-region of a brain.

The discrete equation of dynamic is given by the formula:

$$\sigma_i^{'}=-sign\left(\sum_j \frac{p_j}{\left(s+p\right)N}\left(\sigma_i-\sigma_j\right)-\sum_j \frac{s_j}{\left(s+p\right)N}\left(\sigma_i+\sigma_j\right)+h_i\right).$$

By introducing a weighted majority-minority difference for a system:

$$m_i=\sum_{j\neq i}{\left(s_j+p_j\right)\sigma_j}\Big/{g\left(d\left(i,j\right)\right)\left(s+p\right)},$$

and random parameters to describe effective self-supportiveness of each agent:

$$a_i=\frac{s-p}{s+p}+\frac{\beta}{s+p}s_i\,,$$

we get the dynamical equation in noise absent limit:

$$\sigma_i^{'}=sign\left(m_i\right)\theta\left(\left|m_i\right|-\left|a_i\right|\right)+\sigma_i sign\left(a_i\right)\theta\left(\left|a_i\right|-\left|m_i\right|\right).$$

The mean-field approximation is introduced by replacing the actual value of $m_i$ by its mean value calculated by averaging over disorder values $\langle m_i \rangle$. This equation is valid for slowly decaying interactions, when the equilibrium solution is not reached rapidly.

The further considerations will be performed using the continuous representation of the above discrete equation in a field-theoretical framework. The sum over agents *j* translates in this approach into an integral over *n*-dimensional Euclidean space multiplied by the proper density function. The advantage of mean-field theory is given by the fact that it is able to reduce the full dynamics described by above functional equations with disorder to the averaged functional equation:

$$m'\left(\vec{x}\right) = g\left(\vec{x}, \lfloor m \rfloor\right) + n_0\left(\vec{x}, \lfloor |m| \rfloor\right),$$

provided that $m\left(\vec{x}\right)$ does not change its sign, therefore it is for example close to uniform state [34].

**Continuous limit**

Now we are ready to perform the continuous limit for the dynamic given by the equation (2). In order to search for analytical solutions in the system we constrain ourselves to the case of Euclidean space with arbitrary dimensionality *n*, when only nearest-neighbors couplings are taken into account. Our postulates in this simplified case are presented as follows:

1. *Continuous field* of states

   A new real value field is introduced $v\left(\vec{x}, t\right)$ that describes the state of the system in the point defined by *n*-dimensional vector $\vec{x}$ in the Euclidean space at a given time moment *t*. The field is an abstract representation of a single agent state, allowing for the search of analytical solutions for the system.

2. Positive *strength function*

   The strength of each learning agent is described here using real, positive value, function $f\left(\vec{x}\right)$, i.e. for every vector $\vec{x} \in R^n$, $f\left(\vec{x}\right) > 0$.

3. *Nonlinearity* in the model

   The degree of nonlinearity in the system is governed by the parameter $\beta = 1/kT$, which is introduced in order to ensure the stability of two special states $\pm 1$ that describe to opposite classification outputs of individual learners. This is crucial for machine learning applications of the model, when the binary classification is selected for predicting class membership for new testing data.

4. *Locality* of interactions

   The strength of coupling between neighboring agents is given by the real parameter $\alpha$, and we assume only the nearest-neighbor interactions. This is strongly supported by the Euclidean metric of the space, where fast decaying coupling function $g\left(d\left(i,j\right)\right)$ ensures that only neighbors are connected. In the case of machine learning ensemble of agents ordered for the purpose of the consensus in the Euclidean space by their training parameters values, this assumption means that algorithms with similar values of parameters tend to have similar results. Such observation can be supported by training similar machine learning algorithms on similar data, or can be modeled as strong coupling between neighbors in the Euclidean space of their parameters values space.

Rewriting the equation (4) similarly as in [37], yet for the more general case of *n*-dimensional Euclidean space we find:

$$\sigma_i\left(t+1\right)-\sigma_i\left(t\right)=$$

$$-\sigma_i\left(t\right)+sign\left[\begin{array}{l}\dfrac{s_{i+1}+p_{i+1}}{\left(s+p\right)g\left(d\left(i,i+1\right)\right)}\sigma_{i+1}\left(t\right)+\dfrac{s_{i-1}+p_{i-1}}{\left(s+p\right)g\left(d\left(i,i-1\right)\right)}\sigma_{i-1}\left(t\right)\\ +\dfrac{s_{i+1}-p_{i+1}+s_{i-1}-p_{i-1}}{\left(s+p\right)g\left(d\left(i,i+1\right)\right)}\sigma_i\left(t\right)+2\beta\dfrac{s_i}{\left(s+p\right)}\sigma_i\left(t\right)\end{array}\right],$$

with the continuous limit given by the following substitutions:

$$\sigma_i\left(t+1\right)-\sigma_i\left(t\right)\rightarrow\dot{v}\left(\vec{x},t\right),$$

$$-\sigma_i\left(t\right)\rightarrow -v\left(\vec{x},t\right),$$

$$\frac{s_{i+1}+p_{i+1}}{\left(s+p\right)g\left(d\left(i,i+1\right)\right)}\sigma_{i+1}\left(t\right)\rightarrow f\left(\vec{x}+d\vec{x}\right)v\left(\vec{x}+d\vec{x},t\right),$$

$$\frac{s_{i-1}+p_{i-1}}{\left(s+p\right)g\left(d\left(i,i-1\right)\right)}\sigma_{i-1}\left(t\right)\rightarrow f\left(\vec{x}-d\vec{x}\right)v\left(\vec{x}-d\vec{x},t\right),$$

$$2\beta\frac{s_i}{\left(s+p\right)}\sigma_i\left(t\right)\rightarrow \beta f\left(\vec{x}\right)v\left(\vec{x},t\right).$$

In addition the third term in the *sign* function argument is approaching zero in the continuous limit.

The continuous form of the dynamical equation for the system is therefore given by the formula:

$$\dot{v}\left(\vec{x},t\right)=-v\left(\vec{x},t\right)+f\left(\vec{x}\right)v\left(\vec{x},t\right)-\gamma v^{3}\left(\vec{x},t\right)+\alpha\frac{\hat{\omega}}{\delta\vec{x}}f\left(\vec{x}\right)v\left(\vec{x},t\right). \quad (5)$$

This equation governs the dynamics of learning space, i.e. how the state of agent located in space point $\vec{x}$ and time $t$ is changing during the course of its evolution. The first term describes the process of decaying when there is no coupling to other learners (no self-support). The second one-agent term represents the positive strength function. The third non-linear term weighted by parameter $\gamma$ introduce the global preference for two stationary solutions uniform for the whole system ("YES" and "NO"). The last two-agent term represent the Euclidean metric using only nearest neighbors scaled by $\alpha$ parameter.

The functional description of the model is given by the equation:

$$\dot{w}\left(\vec{x},t\right)=-w\left(\vec{x},t\right)-\gamma\frac{w^{3}\left(\vec{x},t\right)}{f\left(\vec{x}\right)}+\alpha\sqrt{f\left(\vec{x}\right)}\nabla^{2}\sqrt{f\left(\vec{x}\right)}w\left(\vec{x},t\right), \quad (6)$$

with $w\left(\vec{x},t\right)=\sqrt{f\left(\vec{x}\right)}v\left(\vec{x},t\right)$ as a new field. The dynamics of the system is then governed by the general functional form similar to the Schrödinger equation:

$$\frac{\delta H}{\delta\dot{w}\left(\vec{x},t\right)}=\frac{\delta H}{\delta w\left(\vec{x},t\right)}, \quad (7)$$

where H denotes the Hamiltonian (or Lyapunov function) for the system:

$$H=\int\partial\vec{x}\left[\frac{w^{2}\left(\vec{x},t\right)}{2}+\frac{\gamma w^{4}\left(\vec{x},t\right)}{4f\left(\vec{x}\right)}-f\left(\vec{x}\right)\frac{w^{2}\left(\vec{x},t\right)}{2}+\frac{\omega}{2}\nabla\left(\sqrt{f\left(\vec{x}\right)}w\left(\vec{x},t\right)\right)^{2}\right]. \quad (8)$$

Defining the potential energy for the system by

$$V\left(\vec{x},t\right)=\frac{w^{2}\left(\vec{x},t\right)}{2}+\frac{\gamma w^{4}\left(\vec{x},t\right)}{4f\left(\vec{x}\right)}-f\left(x\right)\frac{w^{2}\left(\vec{x},t\right)}{2}+\frac{\omega}{2}\nabla\left(\sqrt{f\left(\vec{x}\right)}w\left(\vec{x},t\right)\right)^{2}, \quad (9)$$

We get more clear form of the dynamic equation:

$$H=\int\partial t\int\partial\vec{x}\left[\frac{\hat{\dot{w}}\left(\vec{x},t\right)}{2}-V\left(\vec{x},t\right)\right]. \quad (10)$$

This form of dynamical equation allows for applying the standard mathematical formalisms of statistical physics, the analytical analysis of the system is therefore much easier in comparison to other types of topology.

**Phases in the system**

The stationary solutions for the system can be computed by using the Thomas-Fermi approximation that neglects the kinetic term in the equation (6), i.e. sets $\alpha = 0$. This approximation and further analytical analysis of the system similar to presented in this manuscript is presented elsewhere [37] for the one dimensional case of the nearest neighbors coupling. We will not repeat here these analysis, yet we would like to recapitulate the generic stationary solutions of the system. We will describe here the together with generic phases that can be observed in dynamics governed by the such general equation (6). The presented here results are valid for any dimensionality of the Euclidean learning space and strongly support the existence of stable solutions for the system governed by real Schrödinger equation similarly to one dimensional case. The solutions for the system are given by the minority clusters surrounded by the majority agents, and the dynamic is of the "staircase" character in the presence of small noise [9, 34, 37].

First, we recapitulate the previous findings by describing three different types of solutions for the equation (6) for different values of parameters $\gamma$ and $\alpha$ [37]:

1) When $\gamma = \alpha = 0$

   The equation of dynamic is given by the equation:

$$\dot{v}\left(\vec{x},t\right) = -v\left(\vec{x},t\right) + f\left(\vec{x}\right)v\left(\vec{x},t\right),$$

   with the stationary solution:

$$v\left(\vec{x},t\right) = e^{\left(\vec{\ }\right)} t v\left(\vec{x},t\right).$$

   The subspace of learning space, where $f\left(\vec{x}\right) > 1$, are not stable, the learners for which $f\left(\vec{x}\right) = 1$ do not change their state, and finally clusters with $f\left(\vec{x}\right) < 1$ in the final state agents does not differ in opinion.

2) When $\gamma > 0$ and $\alpha = 0$

   The stationary solution for a system is given by the equation:

$$f\left(\vec{x}\right) - 1 = \gamma v^2\left(\vec{x},t\right),$$

with two different classes of solutions depending of the sign of $f\left(\vec{x}\right)-1$ term. For larger values of agent strength $f\left(\vec{x}\right)>1$ there are one unstable solution $v\left(\vec{x},t\right)=0$ and two stable ones $v\left(\vec{x},t\right)=\pm\sqrt{\frac{f\left(\vec{x}\right)-1}{\gamma}}$. Learners with self-strength above average influence from others easily get and maintain their state. For weaker agents we have only one stable, stationary solution $v\left(\vec{x},t\right)=0$. Such learners rapidly collapses their state into average consensus value, and cannot maintain their own prediction outcome adjusting themselves to average opinion.

3) When $\gamma<0$ and $\alpha=0$

   The equation of dynamic similarly to one-dimensional case [37] is given by the equation:

$$\dot{v}\left(\vec{x},t\right)=-v\left(\vec{x},t\right)+f\left(\vec{x}\right)v\left(\vec{x},t\right)-\gamma v^{3}\left(\vec{x},t\right),$$

   with the two unstable solutions:

$$v\left(\vec{x},t\right)=\pm\sqrt{\frac{1-f\left(\vec{x}\right)}{-\gamma}},$$

   and one stable one given by the $v\left(\vec{x},t\right)=0$, independent of the actual value of $f\left(\vec{x}\right)$.

Summarizing, three different solutions of the dynamical equation (6) can be observed in the stationary limit. Each agent either support its own prediction outcome, or change its state in accordance with the state of majority of learners. The whole abstract learning space can be divided into subspaces, each with non-zero or zero solution. The clusters with non-zero solution have the mean size proportional to the correlation length in the system.

**Equilibration of the system: adaptive integration procedure**

The equilibration of the system, i.e. the solution close to stationary state, can be described by expanding the original equation (5):

$$\left(f\left(\vec{x}\right)-1\right)v\left(\vec{x}\right)-\gamma v^{3}\left(\vec{x}\right)=-\alpha\frac{\delta}{\delta\vec{x}}f\left(\vec{x}\right)v\left(\vec{x}\right),$$

around the stationary solution from Thomas-Fermi approximation $v_0\left(\vec{x}\right)=\pm\sqrt{\frac{f\left(\vec{x}\right)-1}{\gamma}}$, for $f\left(\vec{x}\right)>1$ or $v_0\left(\vec{x}\right)=0$ otherwise. The dynamical equation is given by the formula:

$$\left(f\left(\vec{x}\right)-1-\gamma v_0^2\left(\vec{x}\right)\right)v\left(\vec{x}\right)=-\alpha\frac{\delta}{\delta\vec{x}}f\left(\vec{x}\right)v\left(\vec{x}\right). \tag{11}$$

This equation describes the changes of states on the border of a selected cluster (where $f\left(\vec{x}\right)\geq 1$). Now, we introduce similarly to one-dimensional case [37] the new variable $w\left(\vec{x}\right)=f\left(\vec{x}\right)v\left(\vec{x}\right)$, namely the state weighted by the strength of the agent, and effective potential $V_{eff}\left(\vec{x}\right)=\left(f\left(\vec{x}\right)-1\right)-\gamma v_0^2\left(\vec{x}\right)=f\left(\vec{x}\right)-1$, for $f\left(\vec{x}\right)>1$ or $V_{eff}\left(\vec{x}\right)=0$, otherwise.

The linear approximation near the cluster edge is therefore described by equation [37]:

$$\frac{f'\left(\vec{x}_0\right)}{f^2\left(\vec{x}_0\right)}\left(\vec{x}-\vec{x}_0\right)w\left(\vec{x}\right)-\gamma v^3\left(\vec{x}\right)=\alpha\frac{\delta}{\delta\vec{x}}w\left(\vec{x}\right),$$

that has solutions as Airy functions for new variable $z={}^{1}\!/_{h}\left(\vec{x}-\vec{x}_0\right)$, where $h^3=\alpha\frac{f^2\left(\vec{x}_0\right)}{f'\left(\vec{x}_0\right)}$. The final stationary solution for the whole cluster given by equation (11) smoothed by the transitory layer with thickness equal to $h$ described by Airy-like function $A_i\left(z\right)\sim z^{-1/4}e^{-2/3 z^{3/2}}$ in the direction perpendicular to the cluster borders in multidimensional space. The thickness $h$ should be much smaller than the average size of the cluster, and the mean distance between clusters.

The statistical description of the whole system can be calculated by averaging a set of variables (such as total area of non-zero clusters, thickness of transitory layer, mean area of one cluster or the number of clusters) over the strength function $f\left(\vec{x}\right)$. For example, the equation for total area of non-zero clusters is given by simple average: $P_{tot} = \theta\left( f\left( \vec{x} \right) - 1 \right)$, the average thickness of transitory layer is given by the formula: $h_{eff} = \langle h \rangle = \left\langle \sqrt[3]{\alpha \frac{f^2\left( \vec{x}_0 \right)}{f'\left( \vec{x}_0 \right)}} \right\rangle$. The order parameter of the system is given by the $\eta = S / 2h_{eff}$, with $S$ as the mean distance between clusters [37]. Three phases of the system can be therefore observed: a) sparse phase $\eta \gg 1$, small minority clusters; b) middle density phase $\eta \geq 1$, some clusters are close to each other; c) large density phase $0 < \eta \leq 1$, with clusters close to each other, and uniformity of opinion is almost reached. The time of collapse of the minority cluster (the equilibration of the system) is given by simple, finite value: $t_{max} = R_0^2 / 2f\left( R_0 \right)$ [37].

The solution for the adaptive integration procedure is given by the minimization of the potential energy for the system (see eq. (9)):

$$\frac{\partial V\left( \vec{x} \right)}{\partial \vec{x}} = \frac{\partial}{\partial \vec{x}} \left[ \frac{w^2\left( \vec{x} \right)}{2} + \frac{\gamma w^4\left( \vec{x} \right)}{4 f\left( \vec{x} \right)} - f(x) \frac{w^2\left( \vec{x} \right)}{2} + \frac{\alpha}{2} \vec{\nabla} \left( \sqrt{f\left( \vec{x} \right)} w\left( \vec{x} \right) \right)^2 \right] = 0,$$

i.e. the solution of the stationary state should be given as rewritten equation (6):

$$-w\left( \vec{x}, t \right) - \gamma \frac{w^3\left( \vec{x}, t \right)}{f\left( \vec{x} \right)} + \alpha \sqrt{f\left( \vec{x} \right)} \nabla^2 \sqrt{f\left( \vec{x} \right)} w\left( \vec{x}, t \right) = 0.$$

The minimization of the global solution should be done numerically in respect to the space variables, with the constrain that both global answers for the system should sum up to one.

We showed above that similarly to the one dimensional case [37] three phases of the system can be observed: sparse (large isolation of agents), middle density (a lot of interesting transient, meta-stable global configurations), and large density state (large value of learning coupling, near the uniformity edge). In the first case clusters of both types of states exist, and when the weak coupling is present there is no bias toward uniform solution. In this regime the intermittent layer approximation is valid and one can estimate the thickness of transient layer and the approximate size of clusters. In the second case a variety of sophisticated geometries, shapes of clusters are present, some are robust and meta-stable, other disappearing slowly changing their state in agreement with majority rule. Here, no analytical solutions are easy to find, therefore computer simulations have to be applied. The selection of the stationary state, i.e. the adaptive integration procedure, can be described by numerical maximization of the free energy of the system with the constrain that the sum of probabilities (the integral over the whole learning space) for both answers should sum up to one.

## Concluding remarks

In this manuscript we have presented the statistical theory of adaptive integration in meta-learning scheme. Summarizing, we would like to stress three major properties that differentiate the model from those studied in our previous paper.

First, in the meta-learning approach only short-range interactions are considered as opposite to Lewenstein et al. [34]. For instance, in the Euclidean two dimensional learning space only nearest-neighbors are coupled. This assumption is well supported by computer vision experiments and simulations, where the information from two dimensional layers of optical sensors is later integrated and analyzed. In the case of sub-regions in a natural brain (like vision cortex), one have to consider much complicated geometries, such as for example cylindrical, three-dimensional learning space. In that case, a network of neurons is integrating information via non-trivial couplings between individual neurons, and the final stationary phase is determined by both topology of connections and strength of couplings between agents. Yet, this process can be locally described as nearest-neighbor coupling in three dimensional Euclidean space.

Second, the random strength parameters simulate different individual features of learning agents, or imperfections of the training data. In the case of matrix of machine learning algorithms, each of them can be described by its intrinsic parameters affecting precision of single classification model of training data. In general case, several different types of machine learning algorithms can be used as individual learners. In that case, the distribution of quality of local prediction can be described as random providing that algorithms differ significantly between each other in terms both of the quality of prediction (classification accuracy), recall values (the ability to memorize the positive items in the training dataset), or precision (the ability to precisely predict the classification of training items).

Third, there are two time scales in the system. The first time scale is related to the fast evolution of individual learners. When input testing data is presented to the system, each learner respond by its own single prediction. This local prediction of each agent is done very rapidly, almost instantly. Then those individual predictions are processed by cellular automata algorithm in order to find the stationary state of the system. This part is denoted as integration of information. As it was shown above, such stationary state has the form of minority clusters surrounded by the sea of majority prediction. Therefore, the final consensus prediction given by the majority rule, still preserves non-orthodox solutions, allowing for fast adaptivity of the system when training data pattern is changed. The time scale for this integrative process is relatively long in comparison to individual predictions, therefore very fast (preferably optimized for parallel processing) cellular automata software implementations have to be prepared in order to apply described above formalism in real life problems. In the statistical model presented here, we assume that there is no coupling between those two time scales. Therefore we neglect all details of individual evolution of learners, focusing our attention for integration phase of incoming local information into the single, consensus answer.

Finally, the coupling between agents is described by the real value impact function of persuasive impact of opposite state agents, relative to supportive impact of the same state learners. The impact strength does not depend on individual features of learners. As it was proposed above we select the $t(x)=x$ , because it will not affect significantly the proposed conclusions.

The dynamic equation (10) addresses the core question of this manuscript, i.e. how typical initial distribution of learners' state evolve in time? Different initial conditions can be distinguished by the numbers of agents sharing opposite opinion $m=\sum_j \sigma_j / N$ into three classes. The first class is close to uniform state $|m|\cong 1$, where almost all learners initially are in agreement. The evolution of the system rapidly collapses into stationary uniform state. This situation is observed when individual learners share similar machine learning algorithms, or wide spectra of parameters values do not change the classification model. Opposite states are sparse, and randomly spread over the learning space. For example, most of single ML algorithms (such as Random Forest, SVM) trained on an easy or with moderate difficulty training dataset will give very similar prediction for a test cases. Therefore the initial state of the consensus system is close to uniformity of opinion, and the uniformity state is the most frequent final state. The second initial condition $0\square$ [illegible] describes much richer solutions space, moderate number of agents share opposite state and those can be distributed randomly over the learning space, or clustered into well defined groups. This describe the situation, where different values of parameters can cause different classification outcomes, or ensemble of ML algorithms contain significantly different between each other algorithms, that construct distinct classification models of input training data. The system has its intrinsic preferences (or in other words preferable local classification model) – most of agents agree with their preferences, yet to some degree the opposite consensus state is possible. Therefore, one can assume that agreement between agents is possible, even if there is significant proportion of learners that classify input data oppositely. The third type of initial conditions $|m|\cong 0$ contain distributed randomly or clustered different agents' states spread over the learning space. Because the number of opposite states is similar, therefore the system is on the edge of the phase transition between two final consensus answers: "YES" or "NO". Therefore, even small perturbation of initial state, parameters change or type and nature of testing examples can in principle guide the system into different, opposite answers. This type of consensus is more fitted to the difficult training cases, where both answers are very probable. Here, the final trained system is very fragile and strongly depend on testing input data. The small change of input testing data can build up the very different consensus value. Here, the consensus as the final, stable state of the whole system is not obvious, and it can take a significant amount of time. The final state can be either randomly distributed negative learners in the majority of positive states, or clustered minorities.

Similar approach to the social influence theory can be applied to other problems, where no obvious coupling can be proposed, yet the consensus between large set of learning agents has to be done. In this way, the mean-field model of Lewenstein et al. and Plewczynski [9, 34, 37] proved that the final solution in the stationary limit exist, and the presence of minority clusters is also observed. Therefore, this approach can be straightforwardly applied in learning agents, or consensus machine learning (meta-learning) problems, where a set of individual learners exchange their prediction results allowing for integration, i.e. common, single opinion formation. The small clusters of minority opinion are still preserved in the ensemble, therefore adaptivity to new training data, or incoming information can be performed much faster. We use the term of *adaptive integration* for denoting this type of consensus, or meta-learning procedure that stores the final answer dynamically (as a stationary state instead of stationary result), preserving minority reports that do not agree with majority rule, therefore allowing for rapid changes of opinions (similarly as in the statistical mechanics of phase transitions).

**Acknowledgements:**

This work was supported by the Polish Ministry of Education and Science (N301 159735) and other financial sources. I would like to thank Prof. M. Lewenstein (ICREA & ICFO, Barcelona, Spain) and Prof. A. Nowak (Psychology Department, University of Warsaw, Warsaw, Poland), together with Prof. M. Niezgodka and dr. F. Rakowski (both from ICM, University of Warsaw, Warsaw, Poland) for stimulating discussions. The manuscript benefits also from the anonymous reviewers' comments, especially in part covering the elucidation of the main goal of the presented work.